\documentclass[runningheads]{llncs}
\usepackage{graphicx}
\usepackage{cite}
\usepackage{amsmath,amssymb,amsfonts}
\usepackage{algorithmic}
\usepackage{graphicx}
\usepackage{textcomp}
\usepackage{adjustbox}
\usepackage{multirow}
\usepackage{bbold}
\usepackage{siunitx}
\usepackage{hyperref}       
\usepackage{xcolor}
\usepackage{sidecap}
\hypersetup{
    colorlinks,
    linkcolor={blue!50!black},
    citecolor={blue!50!black},
    urlcolor={blue!50!black}
}
\usepackage{amsmath, bm}

\usepackage{pifont}
\newcommand{\cmark}{\ding{51}}%
\newcommand{\xmark}{\ding{53}}%

\begin{document}

\title{Class-Incremental Domain Adaptation with Smoothing and Calibration for Surgical Report Generation}

\author{Mengya Xu\inst{1,4 \thanks{Equal contributions.}}\and Mobarakol Islam \inst{2\footnotemark[1]} \and Chwee Ming Lim\inst{3} \and Hongliang Ren\inst{1,4}}
\institute{
Dept. of Biomedical Engineering, National University of Singapore, Singapore and NUSRI Suzhou, China\and
Dept. of Computing, Imperial College London, UK\and
Dept. of Otolaryngology, Singapore General Hospital, Singapore\and
Dept. of Electronic Engineering, The Chinese University of Hong Kong, Hong Kong\\
\email{mengya@u.nus.edu, mobarakol@u.nus.edu, hlren@ieee.org$^*$}
}

\maketitle              
\begin{abstract}

Generating surgical reports aimed at surgical scene understanding in robot-assisted surgery can contribute to documenting entry tasks and post-operative analysis. Despite the impressive outcome, the deep learning model degrades the performance when applied to different domains encountering domain shifts. In addition, there are new instruments and variations in surgical tissues appeared in robotic surgery. In this work, we propose class-incremental domain adaptation (CIDA) with a multi-layer transformer-based model to tackle the new classes and domain shift in the target domain to generate surgical reports during robotic surgery. To adapt incremental classes and extract domain invariant features, a class-incremental (CI) learning method with supervised contrastive (SupCon) loss is incorporated with a feature extractor. To generate caption from the extracted feature, curriculum by one-dimensional gaussian smoothing (CBS) is integrated with a multi-layer transformer-based caption prediction model. CBS smoothes the features embedding using anti-aliasing and helps the model to learn domain invariant features. We also adopt label smoothing (LS) to calibrate prediction probability and obtain better feature representation with both feature extractor and captioning model. The proposed techniques are empirically evaluated by using the datasets of two surgical domains, such as nephrectomy operations and transoral robotic surgery. We observe that domain invariant feature learning and the well-calibrated network improves the surgical report generation performance in both source and target domain under domain shift and unseen classes in the manners of one-shot and few-shot learning. The code is publicly available at https://github.com/XuMengyaAmy/CIDACaptioning.

\end{abstract}
\section{Introduction}
Automatically generating the description for a surgical procedure can free the surgeons from the low-value document entry task, allow them to devote their time to patient-centric tasks, and do post-operative analysis. Image captioning model cannot generalize well to target domain (TD) since existing domain adaptation (DA) approaches such as consistency learning\cite{sahu2020endo}, hard-soft DA\cite{zia2021surgical} are utilized to solve the domain shift problem assuming that the source domain (SD) and the TD share the same class set. However, this assumption is impractical in a surgical environment where the TD often includes novel instrument classes and surgical regions that do not appear in the SD. 

Class-Incremental (CI) learning methods can learn new instruments absent from SD but will fail if there is a domain shift in robotic surgery\cite{castro2018end, kundu2020class}. Cross-Entropy (CE) loss is sensitive to adversarial samples and leads to poor results if the inputs differ from the training data even a bit\cite{gunel2020supervised}. To overcome these issues, supervised contrastive (SupCon) learning\cite{khosla2020supervised} applies extensive augmentation and maximizes the mutual information for different views. In this work, we incorporate SupCon with CI learning for novel TD instrument classes under surgical domain shift.

Most recently, transformer based models are showing state-of-the-art performance in the task of classification \cite{dosovitskiy2020image}, medical image segmentation \cite{chen2021transunet} and caption generation \cite{cornia2020meshed}. The mesh-memory transformer~\cite{cornia2020meshed} ($M^2$ transformer) forms of multi-layer encoder-decoder and learn to describe object interaction using extracted features from the image.
Though the model shows excellent performance in caption prediction, it is unable to deal with domain shift. There is evidence that curriculum by smoothing (CBS)~\cite{sinha2020curriculum} can learn domain invariant features by applying anti-aliasing. Studies have also shown that feature representation can be improved by a well-calibrated model~\cite{islam2020learning} and label smoothing (LS) enhances the model calibration by limiting network from over-confidence prediction~\cite{muller2019does}. In this work, we design class-incremental domain adaption (CIDA) with CI learning and SupCon for novel class adaptation and domain invariant feature extraction. To deal with domain shift and network calibration in the caption generation model, we develop a one-dimensional (1D) CBS and incorporate it with LS for $M^2$ transformer.

Our contributions can be summarized as the following points: (1) Propose CIDA in feature extractor to tackle novel TD classes under domain shift; (2) Design 1D CBS and integrate it with transformer based captioning model to learn domain invariant features and adapt to the new domain for generating the surgical report in robotic surgery; (3) Investigate model calibration with LS for both feature extraction and captioning and observe the effect of well-calibrated model into feature representation and the one/few shot DA; (4) Annotate robot-assisted surgical datasets with proper captions to generate the surgical report for MICCAI robotic instrument segmentation challenge and Transoral robotic surgery (TORS) dataset.

\begin{figure}[!hbpt]
\centering
\includegraphics[width=1\linewidth]{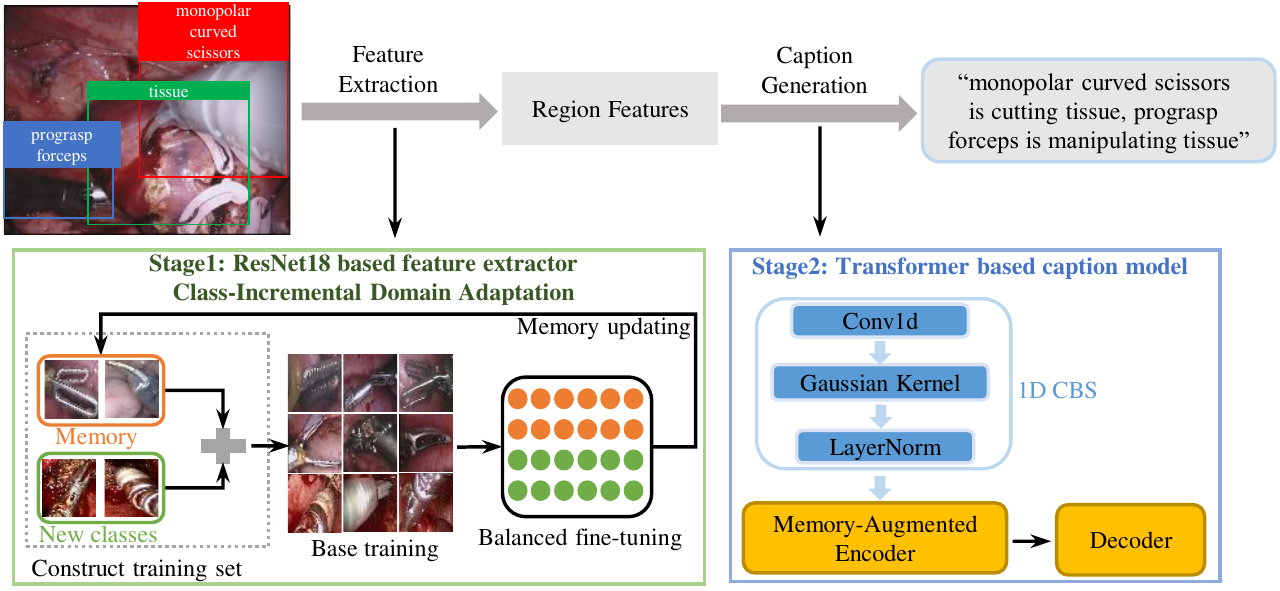}
\caption{Overall workflow. The input image is sent into the ResNet18 based feature extractor augmented with CIDA, and output region features. Inside the transformer-based caption model with 1D CBS, the encoder takes in the region features and understands the relationship between regions. The decoder receives the encoder's output and generates the medical report.}
\label{fig: proposed_model}
\end{figure}

\section{Proposed Method}
Overall caption generation pipeline consists of two stages of networks, as shown in Fig \ref{fig: proposed_model}. The first stage contains a feature extraction model with CIDA and a transformer-based caption model with 1D CBS and LS is designed in the second stage.

\subsection{Preliminaries}
Label Smoothing (LS), where a model is trained on a smoothened version of the true label, $T_{LS} = T(1-\epsilon) + \epsilon/K$, with CE loss, shows great effectiveness in improving model calibration \cite{muller2019does}. The CE loss with LS can be formulate as $CE_{LS} = - \sum_{k=1}^{K} T_{LS} log (P)$ where $T$ is true label, $\epsilon$ is smoothing factor, $K$ is the number of all classes and $P$ is predicted probability.

\subsection{Feature Extraction}
ResNet18\cite{he2016deep} is employed as the feature extractor. CI learning proposed in \cite{castro2018end} aims to handle continually added new classes. However, it is unsuitable to deal with domain shift \cite{kundu2020class} due to the sensitivity of CE loss to training data.

\subsubsection{Class-Incremental learning}
The feature representation and classifier in the feature extractor are updated jointly by minimizing the weighted sum of two-loss functions: CE loss $L_{CE} = - \sum_{k=1}^{K} T log (P)$ to learn the new classes and the distillation loss $L_{DT} = - \sum_{k=1}^{K} T_{dist} log (P_{dist})$ to preserve the learned knowledge from the old classes. $T_{dist}$ and $P_{dist}$ are got by dividing $T$ and $P$ by the distillation parameter $D$\cite{hinton2015distilling}. $D$ is set to 3 for our experiments. 

\subsubsection{Class-Incremental learning with SupCon}
 We decouple the classifier and the representation learning in the feature extraction model and propose a novel CIDA with SupCon loss\cite{khosla2020supervised} to enable the identification of both shared and new classes in the presence of domain shift. The CIDA framework is composed of four steps, as shown in Fig. \ref{fig: proposed_model}. The first step is to construct the training data contains the data from new classes and data from old classes saved in the memory. The second step is the training process which aims to fit a model based on the training data. In the third step, the model is fine-tuned with the balanced subset, consisting of data from memory and partial data from new classes. Each class in the subset has the same number of data. The final step is memory updating which aims to add some data from the new classes into the memory.

\begin{figure}[!hbpt]
\centering
\includegraphics[width=1\linewidth]{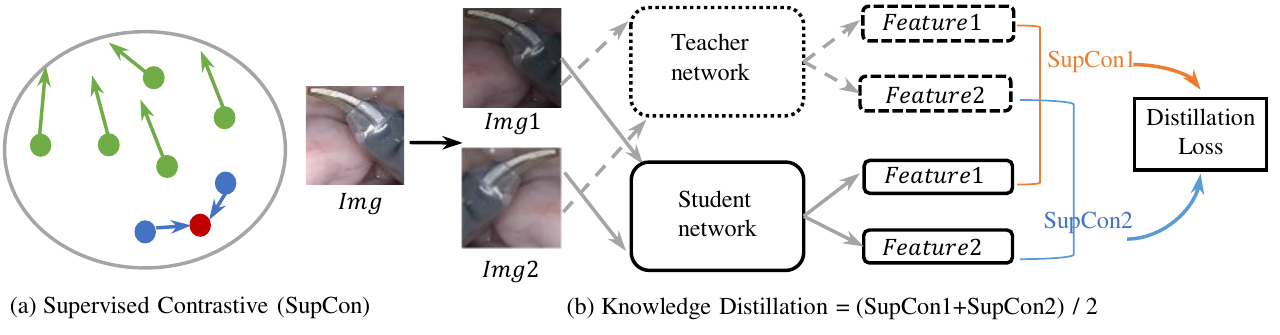}
\caption{Class-Incremental learning with SupCon. (a) explains the idea of SupCon. The red dot represents the original image of one class, blue dots are augmented versions of the same image (Positive images), and the green dots are all the other images in the dataset (Negative images). Positive images are pulled close together and negative images are pushed apart. (b) shows the designed knowledge distillation loss which further upgrades CI method to CIDA method. The student network is the initial copy of teacher network.}
\label{fig:SupCon}
\end{figure}

The feature extractor is trained by minimizing the CI loss consists of SupCon loss\cite{khosla2020supervised} and a novel distillation loss, as shown in Fig. \ref{fig:SupCon}. The SupCon loss function can be formulated as $\mathcal{L}^{s u p}=\sum_{i=1}^{2 N} \mathcal{L}_{i}^{s u p} = \sum_{i=1}^{2 N} ( \frac{-1}{2 N_{\tilde{\boldsymbol{y}}_{i}}-1} \sum_{j=1}^{2 N} \mathbb{1}_{i \neq j} \cdot \mathbb{1}_{\tilde{\boldsymbol{y}}_{i}=\tilde{\boldsymbol{y}}_{j}} \cdot \log \frac{\exp \left(\boldsymbol{z}_{i} \cdot \boldsymbol{z}_{j} / \tau\right)}{\sum_{k=1}^{2 N} \mathbb{1}_{i \neq k} \cdot \exp \left(\boldsymbol{z}_{i} \cdot \boldsymbol{z}_{k} / \tau\right)} )$ which means samples $\boldsymbol{z}_{i}$ and $\boldsymbol{z}_{j}$ which have the same label $\tilde{\boldsymbol{y}}_{i}=\tilde{\boldsymbol{y}}_{j}$ should be maximized in the inner product and pulled together while everything else should be pushed apart in feature representation space~\cite{khosla2020supervised}. Thus the network learns about these random transformations and possesses the potential to handle domain shift.

We design a novel distillation loss for CIDA, which can be constructed using the equation below: 

\begin{equation}
\mathcal L_{DT} =\frac {{L}^{s u p}(Timg1-Simg1) + {L}^{s u p}(Timg2-Simg2)}{2}
\end{equation}

where $Timg1$ represents the feature of Image1 output from the teacher network and $Simg1$ represents the feature of Image1 come from the student network.

\subsubsection{Domain adaptation with CBS}
The $\sigma$, which is the standard deviation of the gaussian kernel, controls the degree of output being blurred after a convolution operation, and increasing $\sigma$ will lead to a greater amount of blur. We implement CBS by annealing $\sigma$, which gradually reduces the amount of blur and allows the model to learn from the incrementally increased information in the feature map. It is also difficult for the model to learn the good representation from the feature maps generated by untrained parameters since these feature maps contain a high amount of aliasing information. Such information can be smoothed out by using a gaussian kernel.

\subsection{Captioning Model}

A transformer-based multi-layer encoder-decoder~\cite{cornia2020meshed} network is used for captioning surgical images. The encoder builds of memory augmented self-attention layer and fully-connected layer. The decoder consists of self-attention on words and cross attention over the outputs of all the encoder layers, similar to \cite{cornia2020meshed}. There are three encoder and decoder blocks stack to encode input features and predict the word class label. The encoder module takes features of the regions from images as input and understands relationships between regions. The decoder reads each encoding layer's output to model each word's probability in the vocabulary and generate the output sentence.

\subsubsection{Feature representation with 1D CBS}
A 1D Gaussian Kernel can be formulated as $g(x) = \frac{1}{\sqrt{2\pi}\sigma}e^{-\frac{x^2}{2\sigma^2}}$, where $g(x)$ represents the x spatial dimensions in the kernel. We propose to augment $M^2$ Transformer\cite{cornia2020meshed} using our designed 1D Gaussian Kernel layer equipped with curriculum learning by annealing the $\sigma$ values as training progresses.
\begin{equation}
y_{i}=\operatorname{ReLU}\left(\operatorname{LayerNorm}\left(\theta_{G_{\sigma}} \circledast\left(\theta_{w} \circledast x_{i}\right)\right)\right)
\end{equation}
where $x_{i}$ is the input region features, $\theta_{w}$ are the learned weights of the 1D convolutional kernel, $\theta_{G_{\sigma}}$ is a Gaussian Kernel whose standard deviation is $\sigma$, $\operatorname{ReLU}$ is a non-linearity\cite{nair2010rectified}, $y_{i}$ is the output of the layer. A single 1D CBS layer is added at the beginning of the encoder (as shown in Fig \ref{fig: proposed_model}). By applying the blur to the output of a 1D convolutional layer, output features are smoothed, and high-frequency information is reduced.

\section{Experiments}
\subsection{Dataset} 
\subsubsection{Source Domain} 
The SD dataset is from the MICCAI robotic instrument segmentation dataset of endoscopic vision challenge 2018 \cite{allan20202018}. The training set includes 15 robotic nephrectomy operations obtained by the da Vinci X or Xi system. 9 objects (instruments and surgical region) appear in the dataset. These instruments have 11 kinds of interaction with the tissue. The captions are annotated by an experienced surgeon in robotic surgery (as shown in Fig. \ref{fig:Dataset_visualization}). The 1st, 5th, 16th sequences are chosen for validation, and the 11 remaining sequences are selected for training following the work\cite{islam2020learning}\cite{xu2021learning}. The splitting strategy ensures that most interactions are presented in both sets.

\subsubsection{Target Domain}
\begin{figure}[!h]
\centering
\includegraphics[width=1\linewidth]{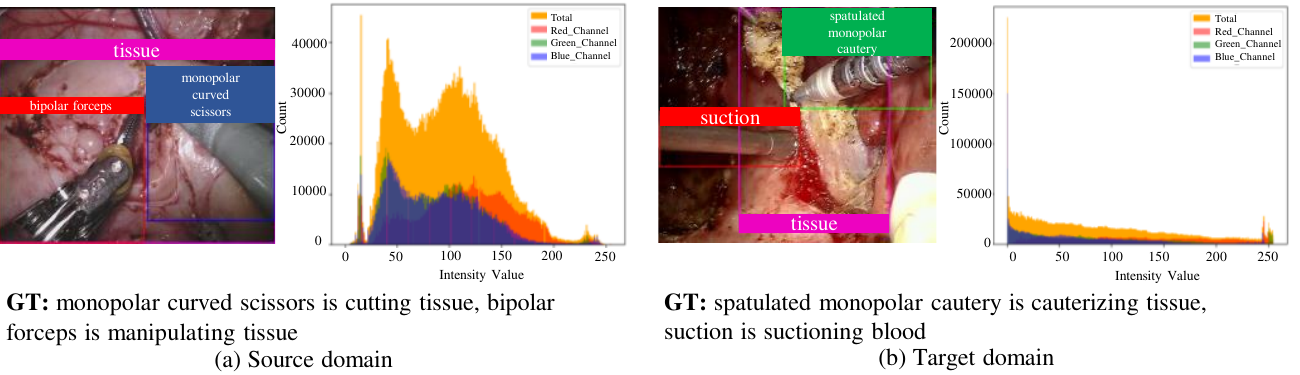}
\caption{Dataset visualization. The frames, captions, and histograms of SD and TD are shown. The histogram can prove the existence of the domain shift.}
\label{fig:Dataset_visualization}
\end{figure}

The TD dataset used for DA is mainly from 11 patients' surgical videos about transoral robotic surgery (TORS) provided by hospitals. The average time length of the video with rich interactions is the 50s. A total of 5 objects and 6 kinds of semantic relationships appear in the dataset. TORS is used for TD experiments where it appears new instruments with different tissues and different surgical backgrounds (Fig. \ref{fig:Dataset_visualization}). The TD dataset is further expanded with frames extracted from one youtube surgical procedure video\footnote{https://youtu.be/bwpEul4KCSc} which is about robotic nephroureterectomy with daVinci Xi.

\subsection{Implementation details}
The feature extractor is trained using stochastic gradient descent with a batch size of 20, weight decay of 0.0001, and momentum of 0.6. We perform 50 and 15 epochs for training and balanced fine-tuning in every incremental step. The learning rate (lr) starts at 0.001 and decays with the factor of 0.8 every 5 epochs. The same lr reduction is also used for fine-tuning except that the initial value is 0.0001.
The captioning model is trained using adam optimizer with a batch size of 50 and a beam size of 5. The training epochs are set to 50. 
For all experiments involved CBS, we use an initial $\sigma$ of 1 and decay the $\sigma$ with a factor of 0.9 every 2 epochs. 
The whole network is implemented by PyTorch and trained in the NVIDIA RTX 2080 Ti GPU.

\section{Results and Evaluation}
We evaluate the model using four metrics for image captioning, namely BLEU-n~\cite{papineni2002bleu}, ROUGE~\cite{lin2004rouge}, METEOR~\cite{banerjee2005meteor}, CIDEr~\cite{vedantam2015cider} and miscalibration with Expected Calibration Error (ECE), Static Calibration Error (SCE), Thresholded Adaptive Calibration Error (TACE) [where, threshold = $10^{-3}$], Brier Score (BS)~\cite{nixon2019measuring, ashukha2020pitfalls}. 

\begin{figure}[!h]
\centering
\includegraphics[width=1\linewidth]{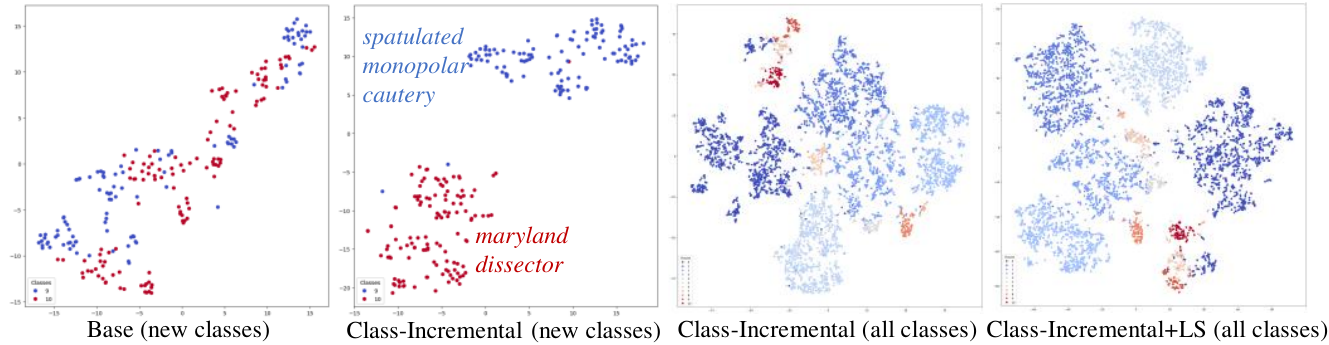}
\caption{tSNEs for two novel classes and all classes. CI learning forms good clusters for the 2 novel instruments. LS leads to tighter clusters.}
\label{fig:tSNE}
\end{figure}

We plot tSNEs for two novel classes and all classes in Fig. \ref{fig:tSNE}. CI learning can extract better features for novel instruments. LS can improve the feature representation in the penultimate layer of the feature extractor. In TD experiments, the first baseline is to first train on the SD and then fine-tune on the TD, and the second baseline is to train directly on the TD. All the proposed methods outperform the baseline in both SD and TD, as shown in Table. \ref{table:final_results}. Class-Incremental learning with SupCon and CBS (CISC) almost obtains the best performance and performs slightly better than Class-Incremental learning with CBS and LS (CICL) in terms of caption metrics and calibration error since SupCon learning can handle the domain shift. The CI method can only deal with novel instruments and DANN can only handle domain shift. SupCon upgrade the CI method to CIDA method which can handle domain shift and novel classes simultaneously, as shown in Table. \ref{table: SupCon_plus_CI_method}. Meanwhile, CBS plays an auxiliary role since it has been proven to achieve better feature extraction\cite{sinha2020curriculum}. Benefit from CBS, even though CICL and CI learning all use CE loss, CICL still performs better than the CI learning method.

\begin{table}[!h]
\caption{Evaluation metrics of the proposed models in SD and TD. The meaning of the abbreviations are: Class-Incremental (CI), Class-Incremental+CBS+LS (CICL), Class-Incremental+SupCon+CBS (CISC), CBS+LS (CL), ResNet18 (Res), $M^2$ transformer (M2T), Domain Adversarial Neural Network (DANN)}
\label{table:final_results}
\resizebox{\columnwidth}{!}{
\begin{tabular}{c|c|c|ccccccc}
\hline
\multirow{3}{*}{Domain} & \multirow{3}{*}{\begin{tabular}[c]{@{}c@{}}Stage 1\\ Feature\\ Extractor\end{tabular}} & \multirow{3}{*}{\begin{tabular}[c]{@{}c@{}}Stage 2\\ Caption\\ Model\end{tabular}} & \multicolumn{7}{c}{\multirow{2}{*}{Metric}} \\
 &  &  & \multicolumn{7}{c}{} \\ \cline{4-10} 
 &  &  & \textbf{BLEU-1}$\uparrow$ & \textbf{BLEU-2}$\uparrow$ & \textbf{BLEU-3}$\uparrow$ & \textbf{BLEU-4}$\uparrow$ & \textbf{METEOR}$\uparrow$ & \textbf{ROUGE}$\uparrow$ & \textbf{CIDEr}$\uparrow$ \\ \hline
\multirow{4}{*}{SD} 
& Res\cite{he2016deep} & X-LAN\cite{pan2020x} & 0.5733 & 0.5053 & 0.4413 & 0.3885 & 0.3484 & 0.5642 & 2.0599 \\ \cline{2-10} 

& Res\cite{he2016deep} & M2T\cite{cornia2020meshed} & 0.5703 & 0.5097 & 0.4572 & 0.4156 & 0.3817 & 0.599 & 2.5385 \\ \cline{2-10} 

& Res\cite{he2016deep} & M2T\cite{cornia2020meshed}+DANN \cite{ganin2016domain} & 0.5995 & 0.5318 & 0.4748 & 0.4301 & \textbf{0.5995} & 0.5994 & 2.4672 \\ \cline{2-10} 
& Res\cite{he2016deep} &  Xu et al\cite{xu2021learning} & 0.5875 & 0.5190 & 0.4599 & 0.4123 & 0.3621 & 0.5982 & 2.5930\\ \cline{2-10} 

 & Res+\textbf{CI} & M2T & 0.5571 & 0.4947 & 0.4395 & 0.3932 & 0.3609 & 0.5791 & 2.319 \\ \cline{2-3}
 & Res+\textbf{CICL} & M2T+\textbf{CL} & 0.6204 & 0.5498 & 0.4923 & 0.4452 & 0.3532 & 0.6017 & 2.6524 \\ \cline{2-3}
 & Res+\textbf{CISC} & M2T+\textbf{CL} & \textbf{0.6246} & \textbf{0.5624} & \textbf{0.5117} & \textbf{0.472} & 0.38 & \textbf{0.6294} & \textbf{2.8548} \\ \hline

  \multirow{4}{*}{\begin{tabular}[c]{@{}c@{}}TD\\ one shot\end{tabular}}
   & Res\cite{he2016deep} & M2T\cite{cornia2020meshed}(direct) &  0.2408 & 0.098 & 0.0319 & 0.     &  0.1051 & 0.2407 & 0.1348\\ \cline{2-10} 
   
  & Res\cite{he2016deep} & M2T\cite{cornia2020meshed}(fine-tune) &  \textbf{0.5678} & 0.4534 & 0.3891 & 0.3305 & 0.2759 & 0.5006 & 1.936\\ \cline{2-10} 
 & Res+\textbf{CI} & M2T & 0.5439 & 0.4568 & 0.403 & 0.352 &  \textbf{0.2886} &  \textbf{0.5279} & 2.2741\\ \cline{2-3}
 & Res+\textbf{CICL} & M2T+\textbf{CL} & 0.5088 & 0.4277 & 0.3758 & 0.3286 & 0.2687 & 0.5071 &  2.3617\\ \cline{2-3}
 & Res+\textbf{CISC} & M2T+\textbf{CL} & \textbf{0.5626} &  \textbf{0.472} &  \textbf{0.417} &  \textbf{0.3648} & 0.2857 & 0.5147 &  \textbf{2.4641}\\ \hline

\multirow{4}{*}{\begin{tabular}[c]{@{}c@{}}TD\\ few shot\end{tabular}}
& Res\cite{he2016deep} & M2T\cite{cornia2020meshed}(direct) & 0.5331 & 0.4567 & 0.4114 & 0.3712 & 0.2738 & 0.5348 & 2.7496 \\ \cline{2-10}

& Res\cite{he2016deep} & M2T\cite{cornia2020meshed}(fine-tune) & 0.5677 & 0.4807 & 0.4285 & 0.3836 & 0.289 & 0.5669 & 2.9209 \\ \cline{2-10} 

& Res\cite{he2016deep} & M2T\cite{cornia2020meshed}+DANN\cite{ganin2016domain} & 0.6338 &0.5367 & 0.4819 & 0.4321 & 0.3173 & 0.5794 & 3.0407
 \\ \cline{2-10}
& Res\cite{he2016deep} &  Xu et al\cite{xu2021learning} &  0.6286 & 0.5422 & 0.4919 & \textbf{0.4457} & 0.3235 & 0.5921 & 3.3620\\ \cline{2-10}

 & Res+\textbf{CI} & M2T & 0.6156 & 0.534 & 0.4859 & 0.4400 & 0.3189 & 0.5975 & 3.2223 \\ \cline{2-3}
 & Res+\textbf{CICL} & M2T+\textbf{CL} & 0.6314 & 0.5434 & 0.4912 & 0.4444 & 0.3262 & 0.6003 & \textbf{3.3930} \\ \cline{2-3}
 & Res+\textbf{CISC} & M2T+\textbf{CL} & \textbf{0.6455} & \textbf{0.5518} & \textbf{0.4935} & 0.4387 & \textbf{0.328} & \textbf{0.6021} & 3.3913 \\ \hline
 
\end{tabular}}
\end{table}

\begin{table}[]
\begin{center}
\caption{Ablation study of the proposed methods}
\label{table: SupCon_plus_CI_method}
\resizebox{\textwidth}{11mm}{
\begin{tabular}{c|c|c|c|c|c|c}
\hline
\multirow{2}{*}{Method} & \multicolumn{3}{c|}{SD} & \multicolumn{3}{c}{TD few shot} \\ \cline{2-7} 
 & \textbf{BLEU1}$\uparrow$ & \textbf{METEOR}$\uparrow$ & \textbf{ROUGE}$\uparrow$ & \textbf{BLEU1}$\uparrow$ & \textbf{METEOR}$\uparrow$ & \textbf{ROUGE}$\uparrow$ \\ \hline
CI & 0.5571 & 0.3609 & 0.5791 & 0.6156 & 0.3189 & 0.5973 \\ \hline
CI+CBS & 0.5704 & 0.3528 & 0.5856 & 0.6185 & 0.3119 & 0.5722 \\ \hline
DANN\cite{ganin2016domain}  & 0.5995 & 0.5995 & 0.5994 & 0.6338 & 0.3173 & 0.5794 \\ \hline
CI+SupCon(our CIDA) & 0.6009 & 0.3963 & 0.6317 & 0.6309 & 0.3205 & 0.6046 \\ \hline
\end{tabular}}
\end{center}
\end{table}

\begin{table}[!h]
\caption{1D CBS. The model trained with 1D CBS is able to learn domain invariant features. Despite some sacrifices in performance on SD, the model achieves excellent performance on TD}
\label{table: CBS_in_captioning_model}
\resizebox{\columnwidth}{!}{
\begin{tabular}{c|c|c|c|c|c|c|c|c|c}
\hline

\multirow{4}{*}{\begin{tabular}[c]{@{}c@{}}Stage 1\\ Feature\\ Extractor\end{tabular}} & \multirow{3}{*}{\begin{tabular}[c]{@{}c@{}}Stage 2\\ Caption\\ Model\end{tabular}} & \multicolumn{4}{c|}{\multirow{3}{*}{SD}} & \multicolumn{4}{c}{\multirow{3}{*}{TD few shot}} \\

 &  & \multicolumn{4}{c|}{} & \multicolumn{4}{c}{} \\
 &  & \multicolumn{4}{c|}{} & \multicolumn{4}{c}{} \\ \cline{2-10} 
 & 1D CBS & \textbf{BLEU-1}$\uparrow$ & \textbf{BLEU-2}$\uparrow$ & \textbf{ROUGE}$\uparrow$ & \textbf{CIDEr}$\uparrow$ & \textbf{BLEU-1}$\uparrow$ & \textbf{BLEU-2}$\uparrow$ & \textbf{ROUGE}$\uparrow$ & \textbf{CIDEr}$\uparrow$ \\ \hline
Res & \xmark & 0.5703 & 0.5097 & 0.599 & 2.5385 & 0.5677 & 0.4807 & 0.5669 & 2.9209 \\ \hline
\multirow{2}{*}{Res+\textbf{CICL}} & \cmark & 0.5972 & 0.5334 & 0.5803 & 2.6499 & 0.6375 & 0.551 & 0.6071 & 3.4956 \\
 & \xmark & 0.5873 & 0.5269 & 0.6067 & 2.9042 & 0.6156 & 0.5261 & 0.591 & 3.2187 \\ \hline
\multirow{2}{*}{Res+\textbf{CISC}} & \cmark & 0.5837 & 0.5264 & 0.6069 & 2.7454 & 0.6478 & 0.559 & 0.6072 & 3.2902 \\
 & \xmark & 0.626 & 0.5651 & 0.6077 & 3.2374 & 0.5669 & 0.4861 & 0.5488 & 2.8877 \\ \hline
\end{tabular}}
\end{table}

\begin{table}[!h]
\caption{Model calibration. LS improves model calibration, boost the SD performance for both two approaches, and improves TD performance for the CISC approach}
\label{tab:calibration}
\resizebox{\columnwidth}{!}{
\begin{tabular}{c|c|c|cc|cc|cccc}
\hline
\multirow{4}{*}{\begin{tabular}[c]{@{}c@{}}Stage 1\\ Feature\\ Extractor\end{tabular}} & \multicolumn{2}{c|}{\multirow{3}{*}{\begin{tabular}[c]{@{}c@{}}Stage 2\\ Caption\\ Model\end{tabular}}} & \multicolumn{2}{c|}{\multirow{3}{*}{SD}} & \multicolumn{2}{c|}{\multirow{3}{*}{TD few shot}} & \multicolumn{4}{c}{\multirow{3}{*}{Calibration Error}} \\
 & \multicolumn{2}{c|}{} & \multicolumn{2}{c|}{} & \multicolumn{2}{c|}{} & \multicolumn{4}{c}{} \\
 & \multicolumn{2}{c|}{} & \multicolumn{2}{c|}{} & \multicolumn{2}{c|}{} & \multicolumn{4}{c}{} \\ \cline{2-11} 
 & 1D CBS & LS & \textbf{BLEU-1}$\uparrow$ & \textbf{CIDEr}$\uparrow$ & \textbf{BLEU-1}$\uparrow$ & \textbf{CIDEr}$\uparrow$ & \textbf{ECE}$\downarrow$ & \textbf{SCE}$\downarrow$ & \textbf{TACE}$\downarrow$ & \textbf{BS}$\downarrow$ \\ \hline
\multirow{2}{*}{Res+\textbf{CICL}} & \multirow{2}{*}{\cmark} & \cmark & 0.6204 & 2.6524 & 0.6314 & 3.393 & 0.1194 & 0.0618 & 0.0613 & 0.6299 \\
 &  & \xmark & 0.5972 & 2.6499 & 0.6375 & 3.4956 & 0.1367 & 0.0627 & 0.0624 & 0.9773 \\ \hline
\multirow{2}{*}{Res+\textbf{CISC}} & \multirow{2}{*}{\cmark} & \cmark & 0.6246 & 2.8548 & 0.6455 & 3.3913 & 0.1385 & 0.0597 & 0.0584 & 0.4346 \\
 &  & \xmark & 0.5837 & 2.7454 & 0.6478 & 3.2902 & 0.1431 & 0.0593 & 0.0592 & 0.9537 \\ \hline
\end{tabular}}
\end{table}

\begin{figure}[!h]
\centering
\includegraphics[width=1\linewidth]{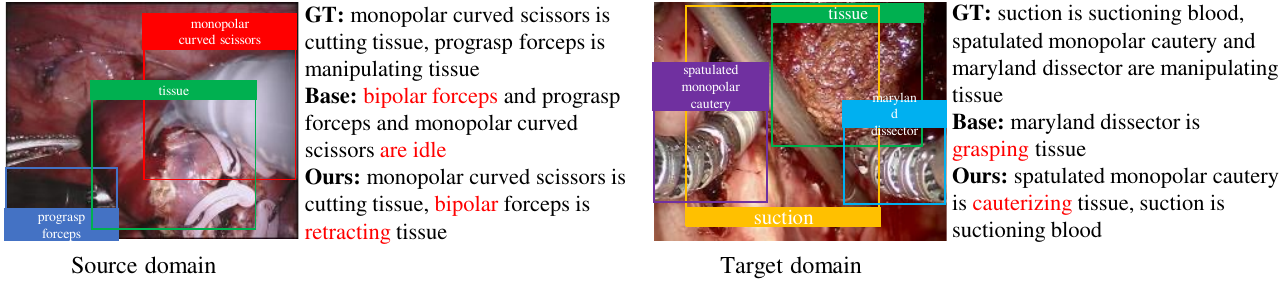}
\caption{The predicted caption of our CIDA method for source and target domain}
\label{fig:predicted_caption}
\end{figure}

Table. \ref{table: CBS_in_captioning_model} investigates the results of applying 1D CBS on the caption prediction model. We observe that 1D CBS enhances the performance in the TD while no improvements in SD. Therefore, 1D CBS helps to learn domain-invariant features. The effects of LS on model calibration and DA are shown in Table. \ref{tab:calibration}. LS can improve the model calibration to fix overconfident prediction and enhances the model robustness and uncertainty. The model trained with LS obtains performance improvement on SD for all the metrics but few metrics for TD. Predicted captions of the CISC approach for SD and few shot TD are shown in Fig. \ref{fig:predicted_caption}. Our model can recognize the instruments and interactions more accurately.

\section{Discussion and Conclusion}
We presented the class-incremental domain adaptation, which aims to handle novel target domain classes under domain shift without the need to re-train all datasets. The feature extractor and transformer-like model trained with CBS can extract the domain-invariant features, and generate the surgical report which describes the instruments-tissue interaction. We also improve the model calibration by using label smoothing. In future work, we will investigate surgical report generation by incorporating temporal information from the surgical video.
\subsection*{Acknowledgements}
This work was supported by the Shun Hing Institute of Advanced Engineering (SHIAE project \#BME-p1-21) at the Chinese University of Hong Kong (CUHK). We would like to express sincere thanks to Lalithkumar Seenivasan for his help on incremental learning of our work.

\bibliography{mybib}{}

\begin{thebibliography}{10}
\providecommand{\url}[1]{\texttt{#1}}
\providecommand{\urlprefix}{URL }
\providecommand{\doi}[1]{https://doi.org/#1}

\bibitem{allan20202018}
Allan, M., Kondo, S., Bodenstedt, S., Leger, S., Kadkhodamohammadi, R., Luengo,
  I., Fuentes, F., Flouty, E., Mohammed, A., Pedersen, M., et~al.: 2018 robotic
  scene segmentation challenge. arXiv preprint arXiv:2001.11190  (2020)

\bibitem{ashukha2020pitfalls}
Ashukha, A., Lyzhov, A., Molchanov, D., Vetrov, D.: Pitfalls of in-domain
  uncertainty estimation and ensembling in deep learning. arXiv preprint
  arXiv:2002.06470  (2020)

\bibitem{banerjee2005meteor}
Banerjee, S., Lavie, A.: Meteor: An automatic metric for mt evaluation with
  improved correlation with human judgments. In: Proceedings of the acl
  workshop on intrinsic and extrinsic evaluation measures for machine
  translation and/or summarization. pp. 65--72 (2005)

\bibitem{castro2018end}
Castro, F.M., Mar{\'\i}n-Jim{\'e}nez, M.J., Guil, N., Schmid, C., Alahari, K.:
  End-to-end incremental learning. In: Proceedings of the European conference
  on computer vision (ECCV). pp. 233--248 (2018)

\bibitem{chen2021transunet}
Chen, J., Lu, Y., Yu, Q., Luo, X., Adeli, E., Wang, Y., Lu, L., Yuille, A.L.,
  Zhou, Y.: Transunet: Transformers make strong encoders for medical image
  segmentation. arXiv preprint arXiv:2102.04306  (2021)

\bibitem{cornia2020meshed}
Cornia, M., Stefanini, M., Baraldi, L., Cucchiara, R.: Meshed-memory
  transformer for image captioning. In: Proceedings of the IEEE/CVF Conference
  on Computer Vision and Pattern Recognition. pp. 10578--10587 (2020)

\bibitem{dosovitskiy2020image}
Dosovitskiy, A., Beyer, L., Kolesnikov, A., Weissenborn, D., Zhai, X.,
  Unterthiner, T., Dehghani, M., Minderer, M., Heigold, G., Gelly, S., et~al.:
  An image is worth 16x16 words: Transformers for image recognition at scale.
  arXiv preprint arXiv:2010.11929  (2020)

\bibitem{ganin2016domain}
Ganin, Y., Ustinova, E., Ajakan, H., Germain, P., Larochelle, H., Laviolette,
  F., Marchand, M., Lempitsky, V.: Domain-adversarial training of neural
  networks. The journal of machine learning research  \textbf{17}(1),
  2096--2030 (2016)

\bibitem{gunel2020supervised}
Gunel, B., Du, J., Conneau, A., Stoyanov, V.: Supervised contrastive learning
  for pre-trained language model fine-tuning. arXiv preprint arXiv:2011.01403
  (2020)

\bibitem{he2016deep}
He, K., Zhang, X., Ren, S., Sun, J.: Deep residual learning for image
  recognition. In: Proceedings of the IEEE conference on computer vision and
  pattern recognition. pp. 770--778 (2016)

\bibitem{hinton2015distilling}
Hinton, G., Vinyals, O., Dean, J.: Distilling the knowledge in a neural
  network. arXiv preprint arXiv:1503.02531  (2015)

\bibitem{islam2020learning}
Islam, M., Seenivasan, L., Ming, L.C., Ren, H.: Learning and reasoning with the
  graph structure representation in robotic surgery. In: International
  Conference on Medical Image Computing and Computer-Assisted Intervention. pp.
  627--636. Springer (2020)

\bibitem{khosla2020supervised}
Khosla, P., Teterwak, P., Wang, C., Sarna, A., Tian, Y., Isola, P., Maschinot,
  A., Liu, C., Krishnan, D.: Supervised contrastive learning. arXiv preprint
  arXiv:2004.11362  (2020)

\bibitem{kundu2020class}
Kundu, J.N., Venkatesh, R.M., Venkat, N., Revanur, A., Babu, R.V.:
  Class-incremental domain adaptation. arXiv preprint arXiv:2008.01389  (2020)

\bibitem{lin2004rouge}
Lin, C.Y.: Rouge: A package for automatic evaluation of summaries. In: Text
  summarization branches out. pp. 74--81 (2004)

\bibitem{muller2019does}
M{\"u}ller, R., Kornblith, S., Hinton, G.E.: When does label smoothing help?
  In: Advances in Neural Information Processing Systems. pp. 4694--4703 (2019)

\bibitem{nair2010rectified}
Nair, V., Hinton, G.E.: Rectified linear units improve restricted boltzmann
  machines. In: Icml (2010)

\bibitem{nixon2019measuring}
Nixon, J., Dusenberry, M.W., Zhang, L., Jerfel, G., Tran, D.: Measuring
  calibration in deep learning. In: CVPR Workshops. pp. 38--41 (2019)

\bibitem{pan2020x}
Pan, Y., Yao, T., Li, Y., Mei, T.: X-linear attention networks for image
  captioning. In: Proceedings of the IEEE/CVF Conference on Computer Vision and
  Pattern Recognition. pp. 10971--10980 (2020)

\bibitem{papineni2002bleu}
Papineni, K., Roukos, S., Ward, T., Zhu, W.J.: Bleu: a method for automatic
  evaluation of machine translation. In: Proceedings of the 40th annual meeting
  of the Association for Computational Linguistics. pp. 311--318 (2002)

\bibitem{sahu2020endo}
Sahu, M., Str{\"o}msd{\"o}rfer, R., Mukhopadhyay, A., Zachow, S.:
  Endo-sim2real: Consistency learning-based domain adaptation for instrument
  segmentation. In: International Conference on Medical Image Computing and
  Computer-Assisted Intervention. pp. 784--794. Springer (2020)

\bibitem{sinha2020curriculum}
Sinha, S., Garg, A., Larochelle, H.: Curriculum by smoothing. arXiv e-prints
  pp. arXiv--2003 (2020)

\bibitem{vedantam2015cider}
Vedantam, R., Lawrence~Zitnick, C., Parikh, D.: Cider: Consensus-based image
  description evaluation. In: Proceedings of the IEEE conference on computer
  vision and pattern recognition. pp. 4566--4575 (2015)

\bibitem{xu2021learning}
Xu, M., Islam, M., Lim, C.M., Ren, H.: Learning domain adaptation with model
  calibration for surgical report generation in robotic surgery. arXiv preprint
  arXiv:2103.17120  (2021)

\bibitem{zia2021surgical}
Zia, A., Bhattacharyya, K., Liu, X., Wang, Z., Kondo, S., Colleoni, E., van
  Amsterdam, B., Hussain, R., Hussain, R., Maier-Hein, L., Stoyanov, D.,
  Speidel, S., Jarc, A.: Surgical visual domain adaptation: Results from the
  miccai 2020 surgvisdom challenge (2021)

\end{thebibliography}
\bibliographystyle{splncs04}
\end{document}